\documentclass[a4paper, 10pt, conference]{IEEEtran}
\IEEEoverridecommandlockouts
\usepackage{cite}
\usepackage{amsmath,amssymb,amsfonts}
\usepackage{graphicx}
\usepackage{textcomp}
\usepackage{xcolor}

\usepackage{times}
\usepackage{caption}
\usepackage{microtype}
\usepackage{booktabs} 
\usepackage{tikz}
\usepackage{float}
\usepackage{algorithm}
\usepackage{algorithmic} 
\usepackage{multicol}
\usepackage{multirow}
\usepackage[bookmarks=true]{hyperref}
\usepackage{subfigure}
\usepackage{enumerate}
\usepackage{amsbsy}
\usepackage{color}
\usepackage{hyperref}
\newcommand{\I}{\textbf{\textit{I}}}
\newcommand{\at}{\textbf{\textit{a}}}

\def\BibTeX{{\rm B\kern-.05em{\sc i\kern-.025em b}\kern-.08em
    T\kern-.1667em\lower.7ex\hbox{E}\kern-.125emX}}
\begin{document}

\title{Graph-Transporter: A Graph-based Learning Method for Goal-Conditioned Deformable Object Rearranging Task\\
\thanks{ ${1}$ Tencent Robotics X Lab, Shenzhen, China. 
{\tt\small \{francisdeng, lipengchen\}@tencent.com}}
\thanks{ ${2}$ The Center for  Intelligent  Control  and  Telescience,  Tsinghua  Shenzhen  International Graduate School, Shenzhen, China 
{\tt\small \{xiachongkun, wang.xq\}@sz.tsinghua.edu.cn}}
\thanks{ ${\dagger}$ Corresponding author}
}
\author{Yuhong Deng$^{1, 2}$, Chongkun Xia$^{2}$, Xueqian Wang$^{2,\dagger}$ and Lipeng Chen$^{1}$} 

\maketitle

\begin{abstract}

Rearranging deformable objects is a long-standing challenge in robotic manipulation for the high dimensionality of configuration space and the complex dynamics of deformable objects. We present a novel framework, Graph-Transporter, for goal-conditioned deformable object rearranging tasks. To tackle the challenge of complex configuration space and dynamics, we represent the configuration space of a deformable object with a graph structure and the graph features are encoded by a graph convolution network. Our framework adopts an architecture based on Fully Convolutional Network (FCN) to output pixel-wise pick-and-place actions from only visual input. Extensive experiments have been conducted to validate the effectiveness of the graph representation of deformable object configuration. The experimental results also demonstrate that our framework is effective and general in handling goal-conditioned deformable object rearranging tasks. 

\end{abstract}

\section{INTRODUCTION}
\label{section:intro}
Although the task of robotic manipulation has been investigated for decades~\cite{throw}\cite{swingbot}\cite{deng2019deep}, most works are focused on rigid objects.Different from rigid object manipulation, deformable objects pose several new challenges. The first challenge is the high dimensionality of the state space~\cite{def_sci}, which leads to the increased complexity of the state observation during deformable object manipulation. The second challenge lies in the complex and non-linear dynamics of deformable materials~\cite{graph_dy}, which makes the state of deformable objects hard to predict under forces or actions.\par
Recently, goal-conditioned deformable object rearranging task has become a research focus, where the robot is required to rearrange a deformable object from an initial configuration to a goal configuration. There are two main solutions for such tasks: \par
\begin{figure}[t]
    \centering
    \includegraphics[width=\linewidth]{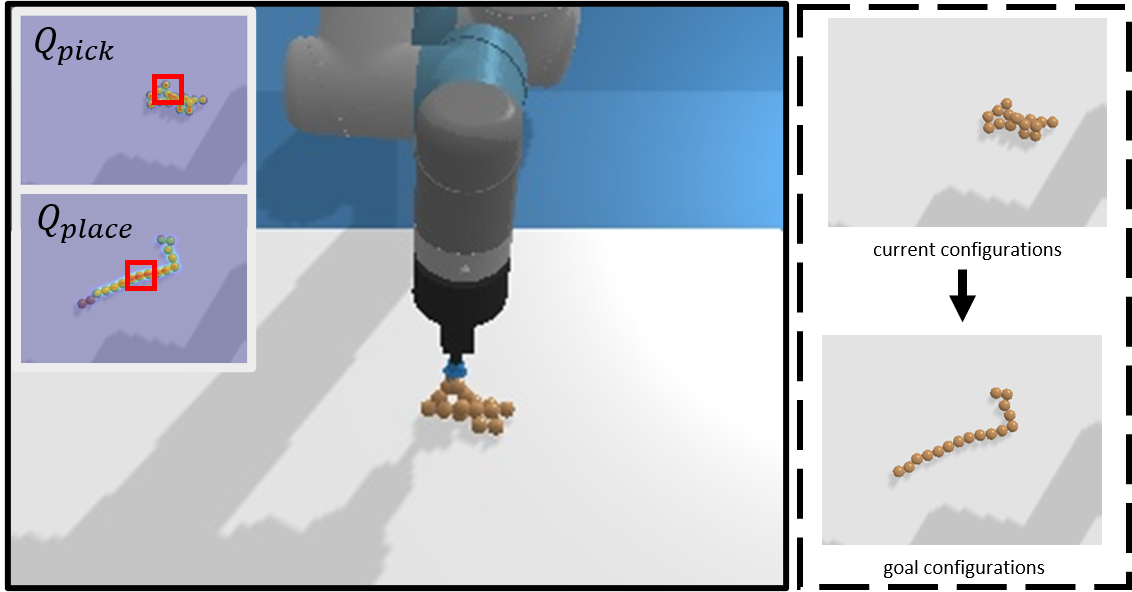}
    \caption{We propose a novel general framework (Graph-Transporter) to output pixel-wise pick-and-place actions from only visual input for goal-conditioned deformable object rearranging tasks. Our framework can learn multi-task policies at the same time.}
    \label{fig:titile}
\end{figure}

The first solution is to establish a data-driven dynamic model~\cite{yan_dynamic} which can predict the resulting object states under certain manipulation actions.  The robot can then determine a proper manipulation action corresponding to the current and goal configurations using sampling-based methods with the dynamic model. However, if the initial and goal configurations of the deformable object differ considerably, a sequence of manipulation actions, rather than a single one, is required for completing the rearranging task. To tackle the multi-step manipulation planning problem, generating and introducing additional sub-goal configurations for the rearranging task has been proved to be an effective strategy~\cite{nair_human}\cite{wang_insert}. However, this line of work can still result in extra effort and introduce accumulative errors.\par

The second solution is to use imitation or reinforcement learning methods to map the raw observations directly to a policy for a specific task~\cite{rl_1}\cite{rl_2} without modeling the object dynamics explicitly. However, this line of work tends to generalize only to a specific set of deformable object manipulation tasks, such as folding cloth, untangling rope-knot~\cite{untangling}, hanging a small towel on a hanger, and relocating cloth and rope in a scene with obstacles~\cite{move}. 
Seita et al.~\cite{transporter_df} proposed a goal-conditioned transporter network, which can be a general learning framework for several goal-conditioned deformable objects rearranging tasks. However, the transport network needs to be retrained for each specific deformable object rearrangement task, which is time-consuming and inconvenient. In contrast, We aim at a general framework that can learn a multi-task policy from demonstrations of different deformable object rearranging tasks together.

To tackle this issue, we first introduce a graph structure to represent the configuration of the deformable object. Compared with the previous CNN-oriented image feature, the graph feature is better at handling the sparse information of the deformable object. It improves the model efficiency and makes it possible to learn policies from demonstration data of a variety of different tasks. We use an unsupervised learning method to detect keypoints from visual input. A representation graph containing vertex and edge features is then established from the keypoints. And we use the representation graph to generate a state vector $\textbf{V}_\text{G}$ by GCN~\cite{gcn} and a mask $\textbf{M}$ related to the position of keypoints. 
We proposed a novel framework, Graph-Transporter, to output pixel-wise actions from only visual observations, where the model learns to process the graph feature from manipulation policy learning. Similar to the transporter network~\cite{transporter, transporter_df}, the output of our framework is the Q-value heatmap and each Q value represents the action confidence at the corresponding pixel position. The mask $\textbf{M}$ is used as the mask and the difference of $\textbf{V}_\text{G}$ of current and goal configurations are used as the convolution kernel in the action Q-value heatmap generation. Our framework is trained in an imitation learning fashion.\par
We have established a dataset and quantitative metrics in simulation and conducted simulation experiments of multiple deformable objects rearranging tasks. The results demonstrate the proposed graph representation method and our framework are effective and general for goal-conditioned deformable object rearranging tasks. The contributions of this paper can be summarized as follows:
\begin{itemize}
\item We propose a method to represent the configurations of deformable objects with graph structure and the model learns to process the graph in the manipulation policy learning.
\item We propose an effective and general learning framework that utilizes graph features and graph convolution for goal-conditioned deformable object rearranging tasks.
\item We build a dataset and quantitative metrics to evaluate the effectiveness of our framework.
\end{itemize}

The paper is organized as follows. The related work is reviewed in Sec.~\ref{section:relatedwork}. The problem is formulated in Sec.~\ref{section:Problem}. Sec.~\ref{section:dataset} establishes our dataset and the proposed framework (Graph-Transporter) is introduced in Sec.~\ref{section:method}. Experimental results are presented in Sec.~\ref{section:experiment}. 

\section{RELATED WORK}
\label{section:relatedwork}
\subsection{Multi-Task Policy of Deformable Object Manipulation}
The application of deformable object manipulation ranges from manufacturing to the service industry~\cite{appli_1}\cite{appli_2}\cite{appli_3}.  Recently, learning and data-driven approaches have paved the way toward equipping robotics with more advanced capabilities in deformable object manipulation~\cite{survey_sci}. Designing a learning policy for a specific deformable object manipulation task has been widely investigated. However, the generalization of these algorithms is limited to specific tasks. Learning multi-task policies for deformable object manipulation has achieved some progress recently. Zeng et al.~\cite{transporter} proposed the transporter network to infer a manipulation sequence from only visual input, which can induce a spatial displacement corresponding to the visual input. Their transporter architecture performs well on several rearranging tasks of rigid objects and Seita et al.\cite{transporter_df} further improved the network and applied it to additional tasks of deformable objects.  Shridhar et al.~\cite{cliport} proposed CLIPort that can produce multi-task policies conditioned on natural language. However, these works must be trained on demonstration data of a specific task to learn a corresponding manipulation policy. We notice that manipulation policies on different rearranging tasks enjoy great similarities. In our framework, the model can learn more general and effective policies that can be used on multiple different deformable object rearranging tasks and the model is trained on the dataset of multiple tasks.

\subsection{{State Representation of Deformable Objects}}
 Considering the high dimensionality of the configuration space of deformable objects, an effective representation method is necessary. Most representation strategies rely on keypoint detection. Early works focus on tracking the keypoints of deformable objects~\cite{tracking_1}\cite{tracking_2}. Miller et al.~\cite{geometric} introduced predefined geometric constraints in the process of detecting feature points to improve detection performance. Some solutions also use CNN feature directly to represent deformable objects, which can result in information redundancy because there are typically no deformable objects in most areas of the image. Ma et al.~\cite{graph_dy} use graph to represent the deformable object and improve modeling accuracy in data-driven dynamic modeling. Since understanding the geometric structure of a deformable object is useful, we group keypoints into a graph structure rather than using them directly, which provides rich semantics including keypoint positions and the topological relation between keypoints. The graph structure is more effective in representing the state of a deformable object by using vertex and edge features. In our model design, the processing of the graph (graph convolution network) is learning from end-to-end manipulation policy learning instead of dynamic modeling separated from manipulation tasks, which can exploit the potential of graphs in manipulation tasks more fully.

\section{PROBLEM FORMULATION}
\label{section:Problem}
Given a current image $\I_{t}$ and a goal image $\I_\text{g}$,  where $t$ denotes the time instant, 
the task of rearranging a goal-conditioned deformable object
is to obtain a manipulation action 
\[
\at_t = \mathcal{F}(\I_\textit{t}, \I_\text{g}),
\]
where $\mathcal{F}$ is the manipulation model that should be developed. This forces the visual image of the deformable object to become as 
\[
\I_{t+1} = \mathcal{D}(\I_\textit{t}, \at_t).
\]
where $\mathcal{D}$ denotes the dynamics of the deformable object.

The above procedure should be repeated until the distance between $\I_\textit{t}$ and $\I_\text{g}$ in the latent space is less than a certain threshold $\alpha$.\par

In this work, our goal is to design an end-end manipulation model $\mathcal{F}$ that can output pixel-wise pick-and-place manipulation actions from only visual input. The challenge of this task mainly lies in two aspects. One is that the target state is not a specific form (a specific shape or a specific location) but a variety of different random forms, which indicates that our model needs to learn more efficient manipulation policies for general rearranging tasks rather than specific policies for a certain rearrangement task. The other is that the task is very complex that requires the model to generate manipulation sequences but without generating any sub-goal images.

\section{DATASET}
\label{section:dataset}
As our goal is to obtain a framework that can learn a general manipulation policy for multiple goal-conditioned deformable object rearranging tasks, we modify and optimize the simulation environment in~\cite{transporter} and come up with our own set of data, which is more convenient for evaluating the generalization and effectiveness of our framework.
The dataset is composed of 1000 deformable object rearranging tasks, while each requires a random number of manipulation actions. In addition, two kinds of deformable objects with different typologies, rope (Fig.~\ref{fig:rope}) and rope ring (Fig.~\ref{fig:rope_ring}), are involved in our task. The established dataset is designed with the explicit goal of training and testing a manipulation policy to rearrange the deformable objects to goal configurations. 

We propose a quantitative definition of the success of a goal-conditioned rearrangement task, which makes our evaluation more informed. The definition is detailed in Sec.~\ref{sec:eva}.

\subsection{Simulation Environment}
Real robotic experiments are usually subjected to a restricted experimental environment and thus not scalable. Safety and cost also limit the number of robot datasets collected in the real environment, and the training of robotic algorithms often requires thousands of iterations. Therefore, we resort to the PyBullet robotic engine~\cite{pybullet}, which can provide satisfying realistic visual renderings and the requirement of deformable object simulation to generate our large-scale demonstration dataset.\par

\begin{figure}[!t]
\centering    
\subfigure[Rope] {
 \label{fig:rope}     
\includegraphics[width=0.45\linewidth]{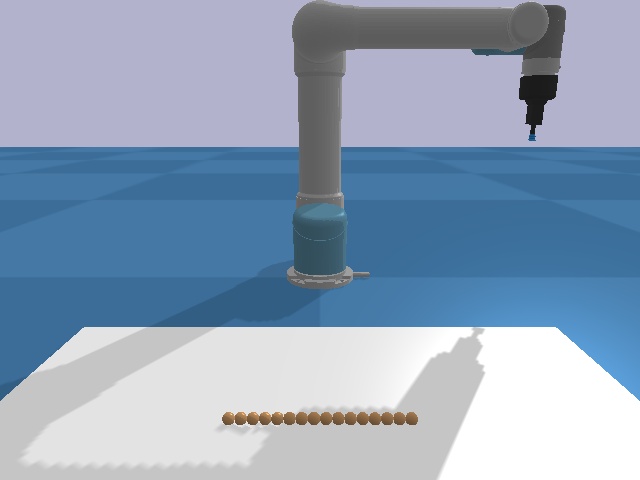}  
}     
\subfigure[Rope ring] { 
\label{fig:rope_ring}     
\includegraphics[width=0.45\linewidth]{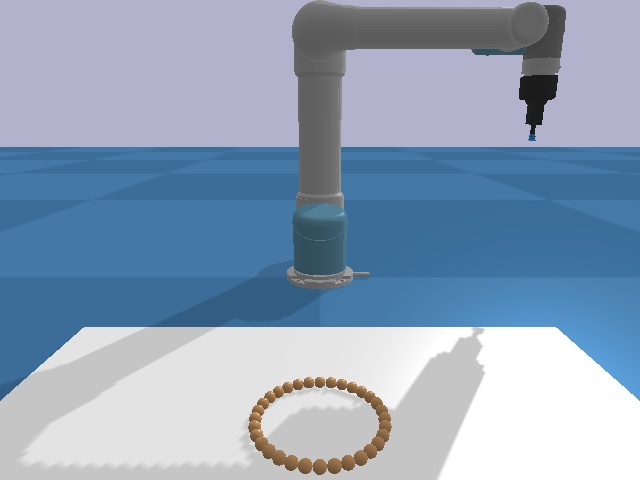}     
}    
\caption{Simulation Environment in PyBullet~\cite{pybullet}: two kinds of deformable objects with different topologies, rope and rope ring, are generated.}     
\label{fig:simulation}     
\end{figure}

For each demonstration trial, one selected deformable object is placed on the platform, and a UR5 robotic manipulator with an end-effector is placed in front of the platform for object manipulation (Fig.\ref{fig:simulation}). Images are captured with a RGB-D camera, which is fixed on the top of the platform.
\subsection{Dataset Generation}
\begin{figure}[!t]
    \centering
    \includegraphics[width=0.75\linewidth]{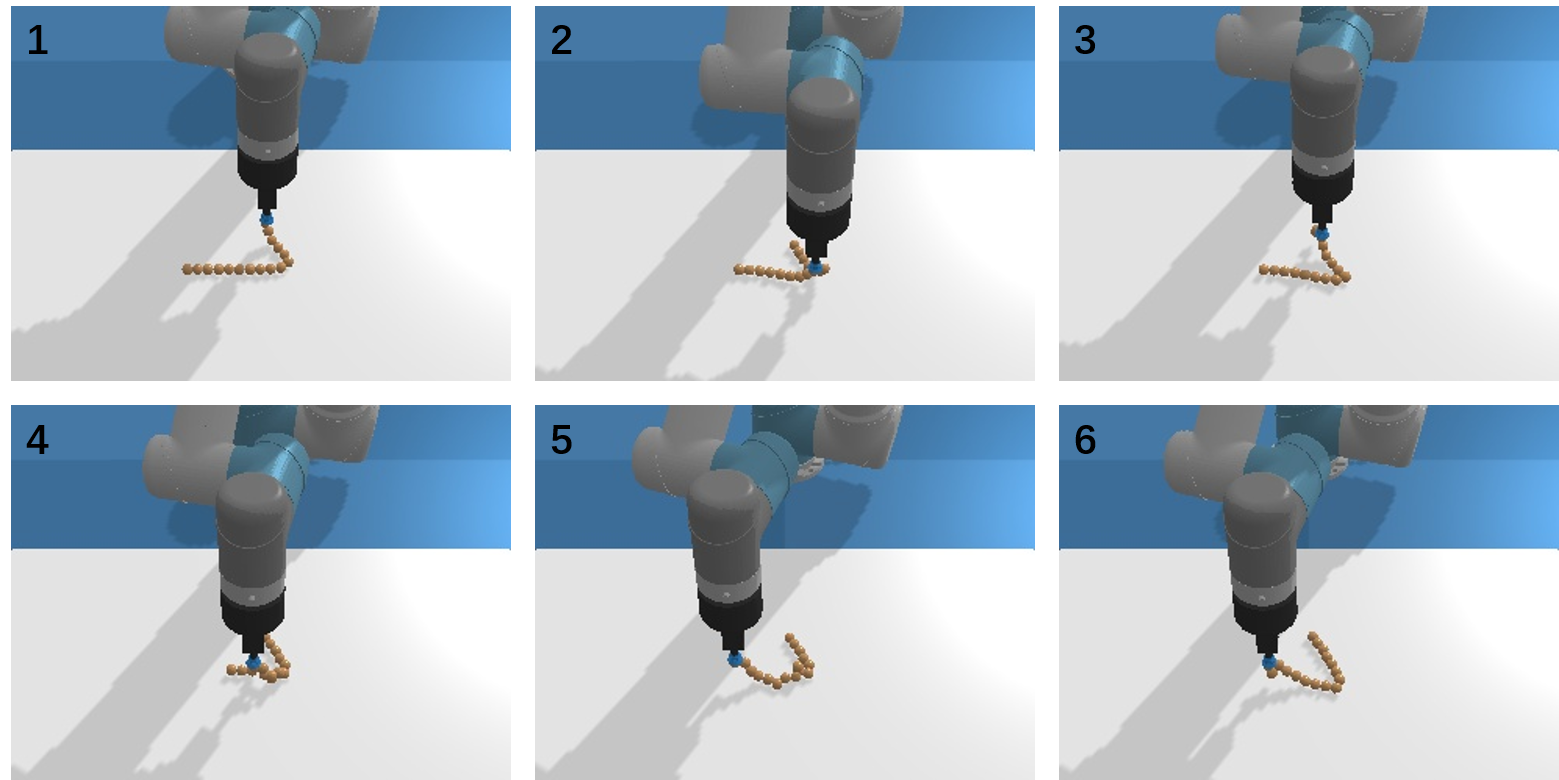}
    \caption{The dataset generation is completed during rearranging the deformable objects guided by an oracle agent (Sec.~\ref{section:imitation}).}
    \label{fig:data_gene}
\end{figure}

We have generated 1000 deformable objects rearranging tasks. For each task, the initial and goal configurations are given as two random images. The goal of the task is to generate a sequence of manipulation actions to rearrange the deformable object to the goal configuration image from the initial image. As shown in Fig.~\ref{fig:data_gene}, The dataset generation is completed during rearranging the deformable objects with manipulation actions produced by an oracle agent (the design of the oracle agent is briefed in Sec.~\ref{section:imitation}). We record and save the demonstration data during the rearranging process. At each time instant $t$, the structure of the record data is $\I_t,\I_g,\at_t\}$, where $\I_t$ is the current image at time $t$, $\I_g$ is the goal image and $\at_t$ is the manipulation action that the robot should take according to $\I_t$ and  $\I_g$. The demonstration data will be collected at every time instant until the task is finished.

\subsection{Dataset Analysis}
\begin{figure}[!b]
    \centering
    \includegraphics[width=0.65\linewidth]{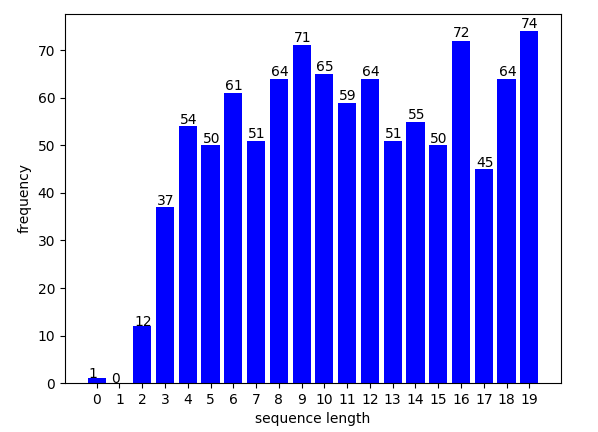}
    \caption{The distribution of sequence length in our dataset.}
    \label{fig:data_dis}
\end{figure}
\begin{figure*}[ht]
    \centering
    \includegraphics[width=0.90\linewidth]{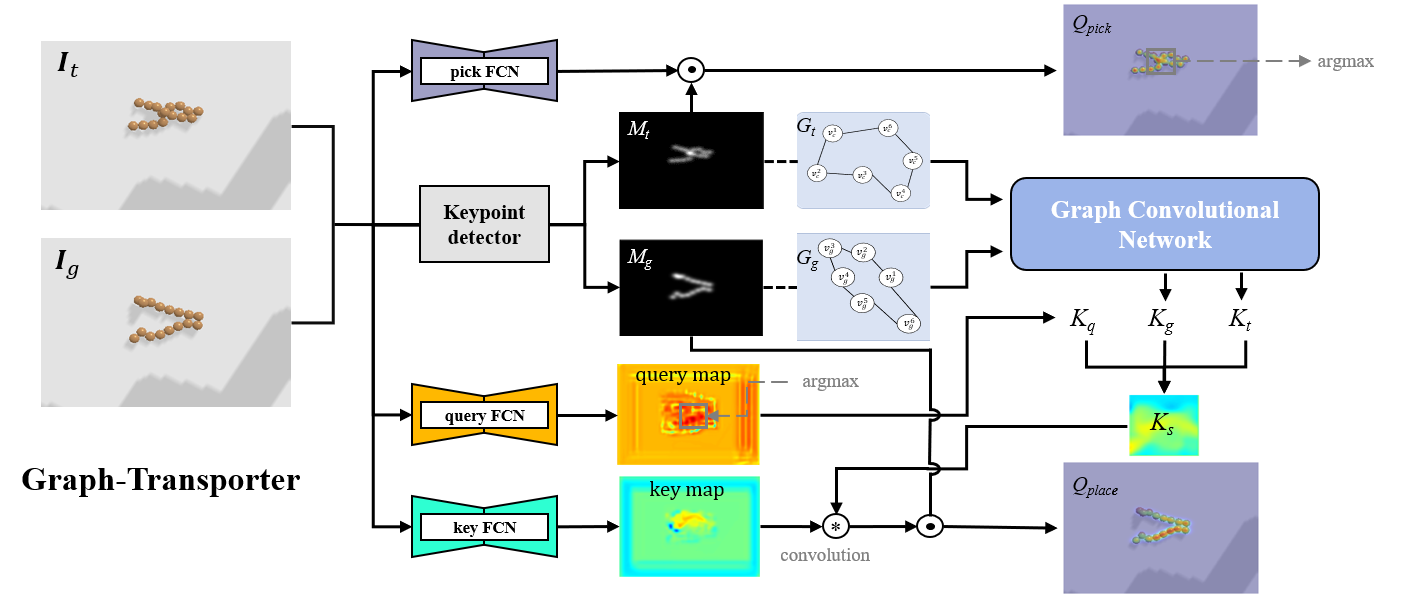}
    \caption{The architecture of Graph-Transporter}
    \label{fig:system}
\end{figure*}

\begin{table}[!t]
    \footnotesize
	\centering
	\caption{Dataset split}
	\setlength{\tabcolsep}{6mm}
	\begin{tabular}{|c|c|c|}
		\hline
		   & rope task & rope-ring task  \\
		\hline
		train & 400 & 400  \\ 
		\hline
		val & 50 & 50  \\
		\hline
		test & 50 & 50  \\
		\hline
	\end{tabular}
	\label{tab:dataset_split}
\end{table}
The constructed dataset is composed of 1000 rearranging tasks, of which the task for rope and task for rope-ring manipulation each account for half. The training set accounts for 80\% of the total data, and the test and eval set accounts for each 10\% of the total data. The spilt of the dataset is shown in TABLE.~\ref{tab:dataset_split}. It is worth noting that our framework can handle multiple goal-conditioned rearranging tasks of rope or rope-ring at the same time without requiring separate training for each specific task.\par

We analyze the distribution of the length of the generated manipulation sequence of all 1000 tasks, which reflects the difficult distribution of tasks in our dataset (the sequence length is equal to the number of manipulation actions required for the corresponding rearranging task). As shown in Fig.~\ref{fig:data_dis}, the sequence length is mainly distributed from 4 to 19, which indicates that our framework must be able to deal with the planning of long sequences and short sequences at the same time. In addition, many tasks require more than 15 manipulation actions to complete, which also illustrates the complexity of the tasks in our dataset.

\section{METHODOLOGY}
\label{section:method}

\subsection{System Overview}
As shown in Fig.~\ref{fig:system}, we introduce Graph-Transporter that utilizes graph features and graph convolution to address the problem of goal-conditioned deformable object rearranging.  We use the pick-and-place actions as the manipulation primitives to complete the task for portability. 

First, we use an unsupervised learning method to extract the keypoints of the deformable object and then construct a graph $G=(V,E)$ to represent the configurations of the deformable object. Then we encode the graph feature to be a state vector by a GCN model. A Gaussian mask related to the positions of keypoints is also generated at the same time. Finally, the graph state vector, Gaussian mask, and RGB images of the current and goal configurations are fed into a FCN-based architecture to generate two Q-value heatmaps for picking and placing. It is worth noting that both Gaussian masks and graph state vectors are generated from RGB images, i.e. our model is an end-to-end formulation from RGB images to manipulation actions.

\subsection{Unsupervised Graph Feature Learning}
The motivation of Graph-Transporter is to use the graph structure to represent the deformable object configuration with a large degree of freedom. Considering that a lot of image sequences can be obtained from our dataset, we use self-supervised learning methods commonly used in image sequences (video) to detect keypoints. We borrow the architecture from~\cite{keypoint} to design our keypoint detector. \par
\begin{figure}[t]
\centering    
\subfigure[Gaussian mask of a rope] {
 \label{fig:mask_rope}     
\includegraphics[width=0.36\linewidth]{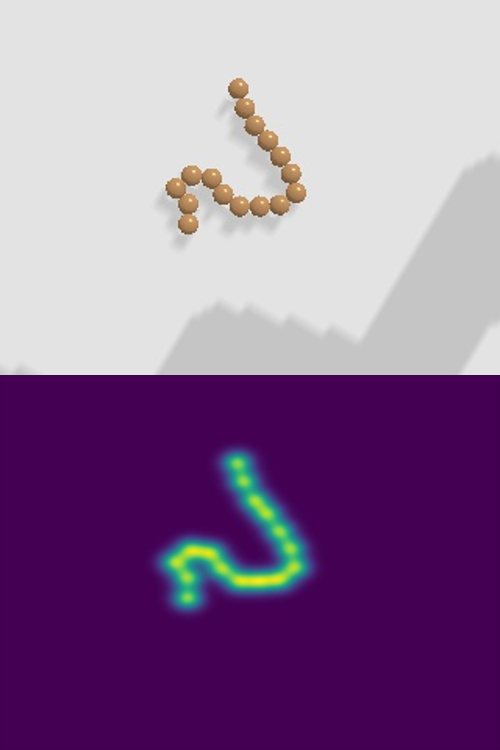}  
}     
\subfigure[Gaussian mask of rope ring] { 
\label{fig:mask_rope_ring}     
\includegraphics[width=0.36\linewidth]{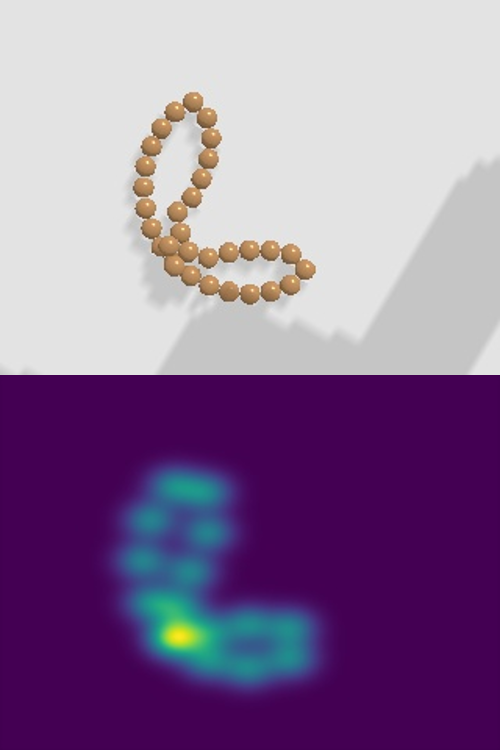}     
}    
\caption{Result of keypoint detection.}     
\label{fig:extact}     
\end{figure}

The key idea of the keypoint detector is that the information of keypoints should be sufficient for the image reconstruction. The model consists of three parts: an encoder $f_e$ to extracting the feature $f_e(\I)$, a point detector $f_{kp}$ to detect keypoints $f_{kp}(\I)$ and generate a Gaussian heatmap $\mathcal{H}_{f_{kp}(I)}$ related to keypoints coordinates, and a decoder $f_{d}$ to reconstruct the image from the extracting feature and Gaussian heatmap. The source image $\I_\textit{src}$ and target image $\I_\textit{tgt}$ (two adjacent images in each demonstration sequence in our dataset) are passed through $f_e$ and $f_{kp}$ to get features and Gaussian heatmap, and then the target image will be reconstructed by the decoder $f_{d}$.
The loss during training is designed as the mean square error of image reconstruction:
\begin{equation}
loss = \left \Vert f_{d}\big( f_e \left( \I_\textit{src} \right), f_e\left( \I_\textit{tgt} \right),\mathcal{H}_{f_{kp}\left( \I_\textit{src} \right)}, \mathcal{H}_{f_{kp}\left( \I_\textit{tgt} \right)} \big), \I_\textit{tgt} \right \Vert
\label{eq:loss}
\end{equation}
 We set the number of keypoints as sixteen, as sixteen keypoints are sufficient to represent a deformable object in our task and can ensure that the noises in the image will not be recognized as keypoints. The example results of keypoints detection are shown in Fig.~\ref{fig:extact}. The Gaussian mask with fixed-variance Gaussian distribution around each keypoint will be also used in action generation.\par

Then we construct a graph $G=(V,E)$ based on the detecting keypoints $f_{kp}(\I)$. Because the position is the most important feature in rearranging tasks, we encoder the position of each keypoint in $f_{kp}(\I)$ as the feature vector of the corresponding vertex in the graph to build up $V$. Then we calculate the adjacency matrix $E$. Considering that a node on a deformable object tends to have a topological constraint relationship with its two adjacent nodes on the left and right, we think each vertex in the graph to hold an adjacency relation with its two nearest vertexes. The elements related to each vertex and its two nearest vertexes are set to be 1 while the other elements are set to be 0 in the adjacency matrix $E$.
\subsection{Model Architecture}
As shown in Fig.~\ref{fig:system}, Graph-Transporter uses three parallel Fully Convolution Networks (FCN), one for picking Q-value heatmap generation and the other two for placing Q-value heatmap generation. First, the current image $\textit{I}_\text{t}$ and the goal image $\textit{I}_\text{g}$ are passed through the point detector, which is pretrained for detecting keypoints by unsupervised learning. Then we can get representation graphs ($G_t$ and $G_g$) and Gaussian masks ($M_t$ and $M_g$) related to the detecting keypoints in $\textit{I}_\text{t}$ and $\textit{I}_\text{g}$. We design a 2-layers GCN model to encode $G_t$ and $G_g$ and the two encoded graph features are fattened into two feature vectors $K_t$ and $K_g$. The calculating process of the GCN layer is defined as:
\begin{equation}
h_i^{l+1} = \sigma\sum_{j\in N_i} \frac{1}{c_{ij}}h_j^lw^l
\label{eq:gcn}
\end{equation}
where $h_i^{l+1}$ represent the features of node $i$ in layer $l+1$, $N_i$ represent all neighbor nodes of $i$ in the graph, $c_{ij}$ is the normalization factor calculated from the adjacency matrix of graph, $w^l$ is the weight of layer $l$, and $\sigma$ represent the activation function.\par
At the same time, the current image $\textit{I}_\text{t}$ and the goal image $\textit{I}_\text{g}$ are concatenated together as input, and then passed through three FCNs in parallel. For the picking point, we multiply the output of the pick FCN by current mask $M_t$ to generate a picking Q-value heatmap ${Q}_{pick}$. For the placing point, we borrowed the attention mechanism, the output of query FCN is treated as a query map while the output of key FCN is treated as a key map. After getting the ${Q}_{pick}$, we can get the picking position. Then we crop an area $K_q$ near the picking position on the query map. A cross-convolution operation is applied on the key map by using the convolution kernel $K_s=K_q+K_g-K_t$. Finally, we multiply the result of cross-convolution by goal mask $M_g$ to generate placing Q-value heatmap ${Q}_{place}$.\par

\begin{algorithm}[!h]
    \caption{imitation policy for rearranging task}
    \begin{algorithmic}[1]
        \STATE unit number = n
        \STATE index = [0,1,..,n-1]
        \STATE bestmap = [0,1,..,n-1]
        \STATE min = $\infty$
        \STATE max = $0$
        \STATE pick = $0$
        \STATE place = $0$
        \FOR {i in index}
        \STATE remap = [i,i+1,..,n-1,0,1,..i-1]
        \STATE dis = MSE($P_c$[remap],$P_g$)
        \IF{dis$<$min}
        \STATE min = dis
        \STATE bestmap= remap
        \ENDIF
        \ENDFOR
        \FOR {j in index}
        \STATE dis1 = MSE($P_c$[bestmap][j],$P_g$[j])
        \IF{dis1$>$max}
        \STATE max = dis1
        \STATE pick= j
        \STATE place= j
        \ENDIF
        \ENDFOR
        \STATE $p_{pick}$ = $P_c$[bestmap][pick]
        \STATE $p_{place}$ = $P_g$[place]
        \STATE \textbf{return} $p_{pick}$, $p_{place}$
    \end{algorithmic}
    \label{alg}
\end{algorithm}

\subsection{Imitation Learning}
\label{section:imitation}
Since we have complete information on the configurations of the deformable object in simulation, we can obtain the best action behavior for the robot to rearrange the deformable object. We adopt an imitation learning methodology to train the robot to mimic the best action behavior.\par
The simulation of deformable objects in PyBullet is realized by discretizing the deformable objects into some small rigid object units that are constrained by each other. We can get the position of every unit during simulation. These position data can be the guidance for the oracle agent to output manipulation action. Denote $P_c$ and $P_g$ are the current and the goal positions of every unit respectively. The generation process of the imitation action $[p_{pick},p_{place}]$ ($p_{pick}$ is the picking position, $p_{place}$ is the placing position) for a rope-ring rearranging task is shown in Alg.~\ref{alg}. First, the algorithm needs to find the node correspondence that minimizes the distance between $P_c$ and $P_g$, which can ensure rotational consistency. After that, we need to find which two corresponding nodes have the largest distance under this node correspondence relationship, and the positions of these two nodes are the pick position and place position. The generation process of imitation action for the rope rearranging task is similar, the only difference is that there are only two types of node corresponding relationships.

\begin{figure*}[ht]
    \centering
    \subfigure[The robot needs to rearrange the shrunk rope into a straight line, whose direction and position are given by the goal image.] {
     \label{fig:kp_rope}     
     \includegraphics[width = 0.90\linewidth]{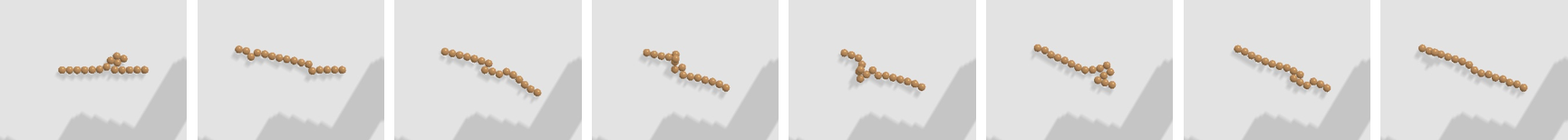}
     } 
    \subfigure[The robot needs to rearrange the rope to a "V" shape, where the angle and direction are given by the goal image.] {
     \label{fig:square}     
     \includegraphics[width = 0.90\linewidth]{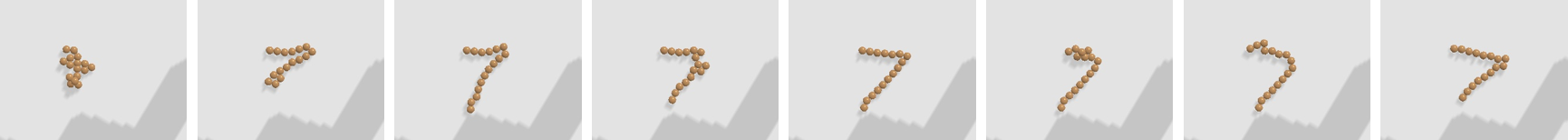}
     }  
    \subfigure[The robot needs to rearrange the rope ring into a square shape, where the right angle is the challenge.] {
     \label{fig:square}     
     \includegraphics[width = 0.90\linewidth]{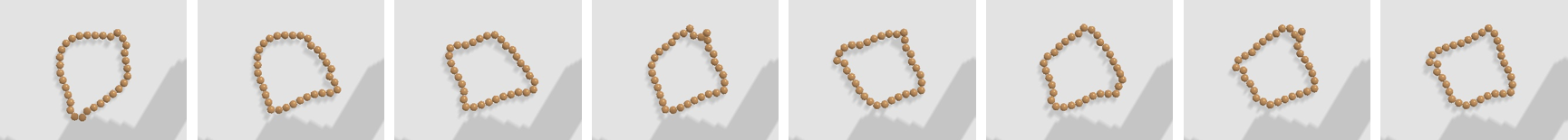}
     }  
     \caption{The experimental results show that our framework can handle multiple deformable object rearranging task without training on each specific task.}
    \label{fig:experi}
\end{figure*}

\section{EXPERIMENTS}
\label{section:experiment}
This section presents experiments to show the performance of our method.

\subsection{Ablation Studies on Graph Feature}
To verify the effect of introducing graph features, we conduct experiments to compare the performance of our model (with graph features) and the goal-conditioned transporter~\cite{transporter_df} (without graph features) on the test data.\par 
The imitated error is the most significant performance metric in imitation learning. Considering that the imitated action is a pixel-wise pick and place action, we use the pixel distance between the picking and placing positions output by the model and the picking and placing positions output by the oracle agent to be the metric to evaluate our model. We also divide the pixel distance by a factor of the image size for normalization. The definition is shown as follows:
\begin{equation}
e_{pick} = \frac{\Vert p_{pick}-o_{pick}\Vert}{\sqrt{w^2+h^2}}
\label{eq:err}
\end{equation}
\begin{equation}
e_{place} = \frac{\Vert p_{place}-o_{place}\Vert}{\sqrt{w^2+h^2}}
\label{eq:err-1}
\end{equation}
where $p_{place}$, $p_{pick}$ are the output actions, $o_{place}$. $o_{pick}$ are the imitated actions. $w$ and $h$ are the width and height of the image respectively.
It should be noticed that we adjust the positions of the camera and the UR5 manipulator to ensure that the camera is directly above the plane where the deformable object is located. So we do not use the visual representation from multi-view image synthesis as used in~\cite{transporter}, but directly use the raw image from the camera (the deformable object is not occluded in the image) in our experimental setup. This setting does not affect the performance of the model but makes experiments faster.\par
The result is shown in TABLE.~\ref{compare_result}. In both tasks, the performance of Graph-Transporter is better than Transporter without graph. The reason is that our graph structure well represents the configurations of the deformable object, which helps the robot learn the rearranging actions better. The mask obtained from the graph structure eliminates invalid regions in the image and speeds up the convergence of the model. At the same time, the convolution kernel constructed by the feature vector obtained by encoding the graph structure brings information about the configuration change of the deformable object for action generation. 

\begin{table}[!t]
  \centering
  \fontsize{7}{8}\selectfont
    \caption{Comparison of two models on two types of deformable object manipulation tasks.}
    \begin{tabular}{cccccc}
    \toprule
        \multicolumn{1}{c} {\multirow{2}{*}{imitation error}}&
        \multicolumn{2}{c}{rope}&  
        \multicolumn{2}{c}{rope ring}\cr
        \cmidrule(lr){2-3}\cmidrule(lr){4-5}
        &$e_{pick}$&$e_{place}$&$e_{pick}$&$e_{place}$\cr
    \midrule
       	Graph-Transporter &0.045&0.031&0.068&0.020\cr
       	Transporter without graph &0.051&0.036&0.085&0.038\cr
    \bottomrule
    \end{tabular}
    \label{compare_result}
\end{table}

\subsection{Evaluation Experiment}
\label{sec:eva}
We also conduct experiments to evaluate the performance of our framework on rearranging tasks. The robot is given random goal configurations by only visual input, and the robot needs to rearrange the deformable object to the goal configurations without any sub-goal input. We define that the robot completes a rearranging task within 20 manipulation actions as a success and the rest as a failure. The situation of completing a rearranging task is that the average pixel distance between the corresponding units in the current and goal configurations is less than 10 (We have found that this threshold is reasonable through most experiments. The two configurations that satisfy this situation are similar enough in the experiment). The success rate of rearranging rope reaches 73\% in 100 testing tasks and the success rate of rearranging the rope ring reaches 68\% in 100 testing tasks. Considering that our task is a highly generalized task, this result can illustrate the effectiveness of our framework. 
The key to the performance of our architecture is to introduce graph features that can better represent the sparse information of deformable objects in the image in the process of Q-value heatmap generation so that the model can learn more general rearranging policies without expanding the size.\par
Three examples of the experiment are shown in Fig.~\ref{fig:experi}. The first task is straightening the crumpled rope. The second task is rearranging the rope into a V-shape. The third task is rearranging the rope ring into a rectangle. All of these tasks are successfully completed in 20 manipulations.

\begin{figure}[t]
\centering    
\subfigure[self-occlusion rope] {
 \label{fig:overlap_rope}     
\includegraphics[width=0.45\linewidth]{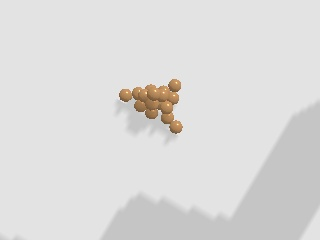}  
}     
\subfigure[self-occlusion rope ring] { 
\label{fig:overlap_ring}     
\includegraphics[width=0.45\linewidth]{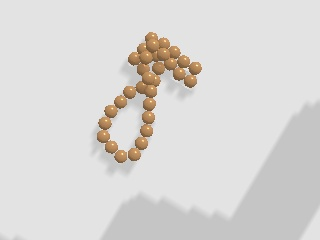}  
}    
\caption{Failure case.}     
\label{fig:fail}     
\end{figure}

\subsection{Failure Case Analysis}
\label{sec:fail}
We have found that the performance of our model will be affected when the deformable object has self-occlusion. The failure case is shown in Fig.~\ref{fig:fail}. It can be difficult to detect effective keypoints when the object has a large degree of self-occlusion, which will make the representation of the graph feature incomplete. At the same time, it will also be hard to differentiate which part of the
deformable object is on the top or on the bottom, which will influence the graph construction and action performance (picking and placing points may be unreachable). We aim at introducing the 3D point cloud to replace the RGB image in future work, where the adjacency relationship in the representation graph will be calculated based on the distance in 3D space instead of pixel distance in the image plane. A 3D graph feature has the potential to fully represent the configurations of a deformable object.
\section{CONCLUSION}
\label{section:conclusion}
In this paper, we propose a  novel framework, Graph-Transporter, to solve the task of goal-conditioned deformable object rearranging. The configurations of the deformable object are represented as a graph in Graph-Transporter. The graph feature is used in a FCN-based architecture, which can generate pixel-wise pick-and-place actions with only visual input. Extensive experiments have been conducted demonstrating the effectiveness of representing the deformable object with graph structure and the rationality of our framework design for deformable object rearranging tasks. In future work, we will extend our framework to more complex deformable objects such as cloth and rubber.



\begin{thebibliography}{10}
\providecommand{\url}[1]{#1}
\csname url@samestyle\endcsname
\providecommand{\newblock}{\relax}
\providecommand{\bibinfo}[2]{#2}
\providecommand{\BIBentrySTDinterwordspacing}{\spaceskip=0pt\relax}
\providecommand{\BIBentryALTinterwordstretchfactor}{4}
\providecommand{\BIBentryALTinterwordspacing}{\spaceskip=\fontdimen2\font plus
\BIBentryALTinterwordstretchfactor\fontdimen3\font minus
  \fontdimen4\font\relax}
\providecommand{\BIBforeignlanguage}[2]{{%
\expandafter\ifx\csname l@#1\endcsname\relax
\typeout{** WARNING: IEEEtran.bst: No hyphenation pattern has been}%
\typeout{** loaded for the language `#1'. Using the pattern for}%
\typeout{** the default language instead.}%
\else
\language=\csname l@#1\endcsname
\fi
#2}}
\providecommand{\BIBdecl}{\relax}
\BIBdecl

\bibitem{throw}
A.~Zeng, S.~Song, J.~Lee, A.~Rodriguez, and T.~Funkhouser, ``Tossingbot:
  Learning to throw arbitrary objects with residual physics,'' \emph{IEEE
  Transactions on Robotics}, vol.~36, no.~4, pp. 1307--1319, 2020.

\bibitem{swingbot}
C.~Wang, S.~Wang, B.~Romero, F.~Veiga, and E.~Adelson, ``Swingbot: Learning
  physical features from in-hand tactile exploration for dynamic swing-up
  manipulation,'' in \emph{2020 IEEE/RSJ International Conference on
  Intelligent Robots and Systems (IROS)}.\hskip 1em plus 0.5em minus
  0.4em\relax IEEE, 2020, pp. 5633--5640.

\bibitem{deng2019deep}
Y.~Deng, X.~Guo, Y.~Wei, K.~Lu, B.~Fang, D.~Guo, H.~Liu, and F.~Sun, ``Deep
  reinforcement learning for robotic pushing and picking in cluttered
  environment,'' in \emph{2019 IEEE/RSJ International Conference on Intelligent
  Robots and Systems (IROS)}.\hskip 1em plus 0.5em minus 0.4em\relax IEEE,
  2019, pp. 619--626.

\bibitem{def_sci}
H.~Yin, A.~Varava, and D.~Kragic, ``Modeling, learning, perception, and control
  methods for deformable object manipulation,'' \emph{Science Robotics},
  vol.~6, no.~54, p. eabd8803, 2021.

\bibitem{graph_dy}
X.~Ma, D.~Hsu, and W.~S. Lee, ``Learning latent graph dynamics for deformable
  object manipulation,'' \emph{arXiv preprint arXiv:2104.12149}, 2021.

\bibitem{yan_dynamic}
W.~Yan, A.~Vangipuram, P.~Abbeel, and L.~Pinto, ``Learning predictive
  representations for deformable objects using contrastive estimation,''
  \emph{arXiv preprint arXiv:2003.05436}, 2020.

\bibitem{nair_human}
A.~Nair, D.~Chen, P.~Agrawal, P.~Isola, P.~Abbeel, J.~Malik, and S.~Levine,
  ``Combining self-supervised learning and imitation for vision-based rope
  manipulation,'' in \emph{2017 IEEE international conference on robotics and
  automation (ICRA)}.\hskip 1em plus 0.5em minus 0.4em\relax IEEE, 2017, pp.
  2146--2153.

\bibitem{wang_insert}
A.~Wang, T.~Kurutach, K.~Liu, P.~Abbeel, and A.~Tamar, ``Learning robotic
  manipulation through visual planning and acting,'' \emph{arXiv preprint
  arXiv:1905.04411}, 2019.

\bibitem{rl_1}
J.~Matas, S.~James, and A.~J. Davison, ``Sim-to-real reinforcement learning for
  deformable object manipulation,'' in \emph{Conference on Robot
  Learning}.\hskip 1em plus 0.5em minus 0.4em\relax PMLR, 2018, pp. 734--743.

\bibitem{rl_2}
Y.~Wu, W.~Yan, T.~Kurutach, L.~Pinto, and P.~Abbeel, ``Learning to manipulate
  deformable objects without demonstrations,'' \emph{arXiv preprint
  arXiv:1910.13439}, 2019.

\bibitem{untangling}
J.~Grannen, P.~Sundaresan, B.~Thananjeyan, J.~Ichnowski, A.~Balakrishna,
  M.~Hwang, V.~Viswanath, M.~Laskey, J.~E. Gonzalez, and K.~Goldberg,
  ``Untangling dense knots by learning task-relevant keypoints,'' \emph{arXiv
  preprint arXiv:2011.04999}, 2020.

\bibitem{move}
D.~McConachie, A.~Dobson, M.~Ruan, and D.~Berenson, ``Manipulating deformable
  objects by interleaving prediction, planning, and control,'' \emph{The
  International Journal of Robotics Research}, vol.~39, no.~8, pp. 957--982,
  2020.

\bibitem{transporter_df}
D.~Seita, P.~Florence, J.~Tompson, E.~Coumans, V.~Sindhwani, K.~Goldberg, and
  A.~Zeng, ``Learning to rearrange deformable cables, fabrics, and bags with
  goal-conditioned transporter networks,'' \emph{arXiv preprint
  arXiv:2012.03385}, 2020.

\bibitem{gcn}
M.~Defferrard, X.~Bresson, and P.~Vandergheynst, ``Convolutional neural
  networks on graphs with fast localized spectral filtering,'' \emph{Advances
  in neural information processing systems}, vol.~29, pp. 3844--3852, 2016.

\bibitem{transporter}
A.~Zeng, P.~Florence, J.~Tompson, S.~Welker, J.~Chien, M.~Attarian,
  T.~Armstrong, I.~Krasin, D.~Duong, V.~Sindhwani \emph{et~al.}, ``Transporter
  networks: Rearranging the visual world for robotic manipulation,''
  \emph{arXiv preprint arXiv:2010.14406}, 2020.

\bibitem{appli_1}
P.~Jim{\'e}nez, ``Survey on model-based manipulation planning of deformable
  objects,'' \emph{Robotics and computer-integrated manufacturing}, vol.~28,
  no.~2, pp. 154--163, 2012.

\bibitem{appli_2}
I.~Leizea, A.~Mendizabal, H.~Alvarez, I.~Aguinaga, D.~Borro, and E.~Sanchez,
  ``Real-time visual tracking of deformable objects in robot-assisted
  surgery,'' \emph{IEEE computer graphics and applications}, vol.~37, no.~1,
  pp. 56--68, 2015.

\bibitem{appli_3}
A.~Kapusta, Z.~Erickson, H.~M. Clever, W.~Yu, C.~K. Liu, G.~Turk, and C.~C.
  Kemp, ``Personalized collaborative plans for robot-assisted dressing via
  optimization and simulation,'' \emph{Autonomous Robots}, vol.~43, no.~8, pp.
  2183--2207, 2019.

\bibitem{survey_sci}
A.~Billard and D.~Kragic, ``Trends and challenges in robot manipulation,''
  \emph{Science}, vol. 364, no. 6446, 2019.

\bibitem{cliport}
M.~Shridhar, L.~Manuelli, and D.~Fox, ``Cliport: What and where pathways for
  robotic manipulation,'' in \emph{Conference on Robot Learning}.\hskip 1em
  plus 0.5em minus 0.4em\relax PMLR, 2022, pp. 894--906.

\bibitem{tracking_1}
T.~Tang, Y.~Fan, H.-C. Lin, and M.~Tomizuka, ``State estimation for deformable
  objects by point registration and dynamic simulation,'' in \emph{2017
  IEEE/RSJ International Conference on Intelligent Robots and Systems
  (IROS)}.\hskip 1em plus 0.5em minus 0.4em\relax IEEE, 2017, pp. 2427--2433.

\bibitem{tracking_2}
T.~Tang and M.~Tomizuka, ``Track deformable objects from point clouds with
  structure preserved registration,'' \emph{The International Journal of
  Robotics Research}, p. 0278364919841431, 2018.

\bibitem{geometric}
S.~Miller, J.~Van Den~Berg, M.~Fritz, T.~Darrell, K.~Goldberg, and P.~Abbeel,
  ``A geometric approach to robotic laundry folding,'' \emph{The International
  Journal of Robotics Research}, vol.~31, no.~2, pp. 249--267, 2012.

\bibitem{pybullet}
E.~Coumans and Y.~Bai, ``Pybullet, a python module for physics simulation for
  games, robotics and machine learning, 2016,'' \emph{URL http://pybullet.
  org}, 2016.

\bibitem{keypoint}
T.~Kulkarni, A.~Gupta, C.~Ionescu, S.~Borgeaud, M.~Reynolds, A.~Zisserman, and
  V.~Mnih, ``Unsupervised learning of object keypoints for perception and
  control,'' \emph{Advances in neural information processing systems}, vol.~32,
  pp. 10\,724--10\,734, 2019.

\end{thebibliography}
\end{document}